# Using the Pepper Robot to Support Sign Language Communication

Giulia Botta[1], Marco Botta[2], Cristina Gena[2][1], Alessandro Mazzei[2], Massimo Donini[2], and Alberto Lillo

[1] Politecnico di Torino, Torino, Italy
[2] Dipartimento di Informatica, Università di Torino, Italy

Abstract. Social robots are increasingly experimented in public and assistive settings, but their accessibility for Deaf users remains quite underexplored. Italian Sign Language (LIS) is a fully-fledged natural language that relies on complex manual and non-manual components. Enabling robots to communicate using LIS could foster more inclusive human–robot interaction, especially in social environments such as hospitals, airports, or educational settings. This study investigates whether a commercial social robot, Pepper, can produce intelligible LIS signs and short signed LIS sentences. With the help of a Deaf student and his interpreter, an expert in LIS, we co-designed and implemented 52 LIS signs on Pepper using either manual animation techniques or a MATLABbased inverse kinematics solver. We conducted a exploratory user study involving 12 participants proficient in LIS, both Deaf and hearing. Participants completed a questionnaire featuring 15 single-choice video-based sign recognition tasks and 2 open-ended questions on short signed sentences. Results shows that the majority of isolated signs were recognized correctly, although full sentence recognition was significantly lower due to Pepper's limited articulation and temporal constraints. Our findings demonstrate that even commercially available social robots like Pepper can perform a subset of LIS signs intelligibly, offering some opportunities for a more inclusive interaction design. Future developments should address multi-modal enhancements (e.g., screen-based support or expressive avatars) and involve Deaf users in participatory design to refine robot expressivity and usability.

## 1 Introduction

Over the past decades, the field of robotics has experienced rapid growth, with robots initially deployed in industrial contexts, for example, as robotic arms in assembly lines or as autonomous vehicles in warehouses [30]. These systems, designed to assist with repetitive or physically demanding tasks, are typically characterized as autonomous or automated machines with high practical utility. However, their interaction with humans has traditionally been limited and functionally driven, differing substantially from human-to-human communication.

---

[1] Corresponding author

Social robots [12], [8], [15] are designed to engage with humans on a social level. They convey communicative intent both verbally and non-verbally, recognize and express emotions, and support users, including those with impairments, in everyday social interactions. As social robots become more capable and accessible, they are expected to emerge as one of the most disruptive technologies of the near future. The increasing availability of commercial social robots, such as Pepper, makes it plausible to envision their widespread adoption in homes and public settings alike [20], [31].

Currently, most social robots interact using speech synthesis and visual displays. However, their integration into inclusive environments raises some question about accessibility. In particular, individuals who are Deaf[2] or hard of hearing continue to face significant communication barriers, as robots or avatars capable of signing remain relatively rare, and text transcription offers only a partial solution. Addressing this gap is a crucial step toward ensuring equitable access to human robot interaction.

In this study, we explore whether, and to what extent, a commercial social robot like *Pepper* can produce *Italian Sign Language* (Lingua Italiana dei Segni, LIS) to support a more accessible interaction. Given Pepper's physical limitations, such as the inability to move its fingers independently, our goal is not to implement the full LIS lexicon but to assess which signs can be feasibly reproduced and to evaluate their intelligibility for LIS users. We aim to address the following research question (RQ1): *What are the technical and communicative limits of using a commercial social robot such as Pepper to express LIS for a more accessible Deaf–robot interaction?*. To achieve this goal, with the help of a Deaf student and his tutor, an expert in LIS, we co-designed and implemented a set of LIS signs on Pepper using both manual animation techniques and a semi-automated method based on inverse kinematics. We then conducted a user study with 12 participants proficient in LIS, participants to assess recognition of individual signs and short signed sentences. This exploratory investigation seeks to contribute to the broader discussion on inclusive robotics and the potential for sign language as a modality in human–robot communication.

The remainder of the paper is structured as follows. Section 2 introduces key concepts related to sign languages and reviews related work. Section 3 describes our implementation of LIS signs on the Pepper robot. Section 4 presents the experimental evaluation and results. Finally, Section 5 offers conclusions and directions for future work.

## 2     Background and Related Works

We begin by briefly introducing the main features of sign languages, followed by a review of relevant work related to our project.

---

[2] We follow the common convention of capitalizing *Deaf* to denote native signers.

## 2.1 Sign Languages

Sign languages are fully natural languages in every respect [9]. They do not have a strict correspondence with specific spoken languages but are governed by grammars that share many lexical and syntactic mechanisms with spoken languages.

Unlike vocal languages, sign languages rely on multiple *articulators*, including hands, facial expressions, and body movements, rather than vocal sound. Signs differ from the spontaneous gestures people use while speaking, as they serve a distinct morpho-syntactic-semantic linguistic function. Each sign language is typically associated with a specific community, often national but sometimes regional. Since sign languages develop independently within each community, the same concept may be expressed using different signs across languages, and conversely, the same sign may convey different meanings [28], [26].

Until recently, sign languages were not officially recognized as full-fledged languages. However, in the past few decades, awareness and recognition of the Deaf community and its languages have grown, accompanied by initiatives to support their use and study. For instance, in 1987, the first official description of LIS was published, and more recently, a comprehensive work has further defined its structure and grammar [6].

In this paper, we focus specifically on LIS and its vocabulary. LIS signs are characterized by several parameters, including location, hand shape, movement, orientation, facial expressions, mouth movements, gaze, and posture. LIS also follows grammatical rules for the formation of complex nouns and verbs and typically adopts a Subject–Object–Verb (SOV) word order for declarative sentences [6].

We are aware of the morpho-syntactic role played by the collocation in the space of the signs and its importance for the correctness and the naturalness of a signed sentence. Similarly, we know the mobility limitations of the Pepper's arms and fingers. Our intention is to provide a preliminary study that shed light on the current possibilities and limitations of a widespread social robot in producing comprehensible signs. We believe that this kind of study can guide and encourage towards the design and the realization of more accessible social robots.

## 2.2 Related Works

Research on the use of social robots for sign language communication began in 2012, with a study that used the NAO H25 robot to teach sign language to Deaf children [16]. The researchers focused on Turkish Sign Language (TSL), comparing the robot's signing to that of a human teacher. The evaluation revealed limitations in the robot's structure and mobility. In subsequent work, the same group used a modified version of the Robovie R3 robot, which could move each finger independently. R3 outperformed NAO in the authors' evaluation of sign production, as its independently articulated fingers allowed for a closer approximation of human hand movements.

In 2019, a study on Persian Sign Language (PSL) introduced a custom-built robot called RASA (Robot Assistant for Social Aims) [25]. The robot featured a human-like face and two hands with independently movable fingers, and was programmed with 70 signs selected to represent diverse PSL configurations. Although the robot executed the signs quite well, participants noted that the lack of facial expressions and mouth movements negatively affected sign clarity and expressiveness.

Another 2019 study focused on teaching sign language to autistic children [2]. Researchers customized the InMoov robot and programmed it with nine signs. The study found that autistic children were engaged by the robot and showed a willingness to imitate its movements, suggesting potential for learning.

Even in 2019, researchers developed a system to convert natural language into Spanish Sign Language (LSE), using the assistive robot TEO [13].

Several recent works have addressed sign language production using virtual avatars focusing on multi-articulator control and linguistic expressiveness. SignAvatar [17] introduces a transformer-based framework capable of generating accurate word-level signs in 3D, capturing realistic hand, face, and body motions. SignAvatars [1] provides the largest dataset of full-body sign language motion, enabling end-to-end training for both generation and recognition tasks. SGNify [4] reconstructs expressive 3D avatars from monocular video using linguistic priors, improving articulation accuracy, particularly in facial expressions and hand configurations. While these systems primarily target American Sign Language (ASL) or British Sign Language (BSL), their multimodal architectures are language independent and can be adapted to LIS.

Many recent studies have focused on multimodal sign language production, aiming to generate coherent and expressive signing sequences by integrating different input sources such as spoken language, textual transcriptions, or gloss annotations[3]. MS2SL [32] investigates how continuous sign language sequences can be generated from either speech or text, emphasizing the alignment between modalities. The approach is evaluated on PHOENIX14T and How2Sign: PHOENIX14T is a benchmark dataset for German Sign Language (DGS) containing parallel sequences of spoken German, gloss annotations and signing videos, while How2Sign includes full-body signing aligned with English instructional videos, offering rich multimodal data. Neural Sign Actors [5] aims to produce realistic full-body signing animations from textual input, focusing on accurate timing and coordination across articulators (hands, face, torso). Mixed SIGNals [10] explores how sign language movements can be composed from reusable motion segments, enhancing the fluidity and expressiveness of the generated signs. A complementary line of research has explored inclusion in cultural settings by developing the GAMGame, a web app designed for the d/Deaf community, which uses affective-driven storytelling and emotion-based recommendations to foster diverse and empathic interpretations of museum content, integrating usergenerated stories with AI-based reasoning systems for cultural sensemaking [18].

---

[3] Gloss annotations are a common convention used to represent sign language utterances in written form by mapping individual signs to approximate equivalents in a spoken language, typically using capitalized words (e.g., HOUSE, GO, I). [11]

Finally, a few research projects specifically studied automatic translation and/or generation of LIS. The ATLAS-LIS4ALL projectc concerned the production from an avatar of LIS sentence regarding weather-forecast and rail-station messages respectively [24], [3], [22]. More recently, a deep learning system for automatic recognition of LIS signs in the news domain has been proposed [21].

## 3    Methodology

The approach adopted in this study was grounded in a participatory and iterative design process, involving members of the Deaf community from the early stages of development. Initially, a preliminary selection of LIS signs was implemented on Pepper through consultation of the *Radutzky* LIS dictionary *"Dizionario bilingue elementare della lingua dei segni italiana"* [27]. However, recognizing the limitations of a top-down design approach, we subsequently involved a Deaf student from the Department of Computer Science and his LIS interpreter in the project. After presenting the project objectives, along with the capabilities and constraints of the Pepper robot, we collaboratively selected a subset of signs and short sentences considered compatible with Pepper's physical affordances (e.g., lack of independent finger movement, limited wrist rotation, restricted handto-body contact, etc.). Each proposed sign was implemented on the robot and demonstrated to the student and his tutor, then iteratively redesigned with their input until they were satisfied. Their feedback informed multiple cycles of refinement and re-implementation. This co-design phase lasted approximately two months.

The initial implementations were created manually using the Pepper SDK's Animation Editor. In a second phase, a semi-automated pipeline was developed to facilitate sign generation, based on a MATLAB inverse kinematics solver and a Python-based exporter to generate Pepper-compatible animation files. This hybrid methodology ensured both linguistic validity and technical scalability. These approaches will be described in the following.

### 3.1    Apparatus and material

Pepper[4] is one of the most used social robotics platform [19]. It features a humanlike head, a built-in touchscreen tablet on its chest, a wheeled base that mimics human legs, and two arms ending in hands with five fingers each. However, the fingers are not independently controllable, they can only open and close simultaneously. This physical limitation makes it impossible to reproduce sign language configurations that require differentiated finger positioning. Consequently, our work focused on signs that can be performed using hand configurations in which all fingers are either open or closed.

---

[4] https://us.softbankrobotics.com/pepper/

Another distinctive feature of Pepper, the chest-mounted tablet, can facilitate interaction in many applications. However, in the context of sign language, it may pose a constraint: the tablet interferes with arm movements, making it difficult or impossible to perform signs that require gestures in close proximity to the torso.

Despite these limitations, we successfully co-designed and implemented a total of 52 LIS signs[5]. The selected signs span several semantic categories, including school-related terms, common actions and nouns, and work-related vocabulary. These signs were primarily created using the Animation Editor, a tool included in the Android Studio plugin Pepper SDK. This tool enables the definition of keyframe animations by specifying joint configurations at individual timesteps, allowing the creation of time-based movement sequences that the robot can execute.

After generating the individual signs, we addressed a more complex challenge: the composition of short sentences. We co-designed and implemented four simple sentences, adhering to the grammatical structure of LIS, which typically follows a Subject–Object–Verb (SOV) word order, as opposed to the Subject–Verb–Object (SVO) order of standard Italian. Each sentence consists of a subject (first person, Pepper), a verb representing an action (e.g., "to eat", "to go"), and an object (or complement).

The following figures illustrate a selection of example signs implemented on the robot: "Amare" (To Love). This sign is performed with both arms, which execute the same movement. The sign consists of a single motion that is repeated twice. The movement begins as shown in Fig.1a and ends as in Fig.1b.

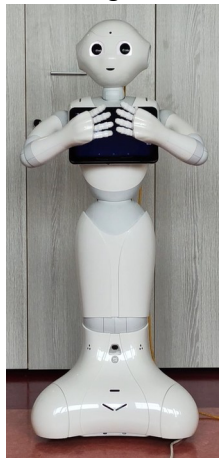  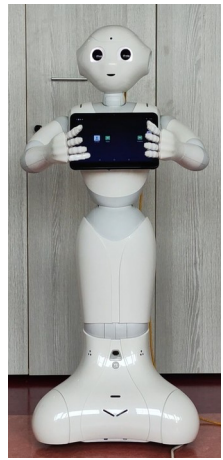

(a) Starting position                    (b) Ending position

---

[5] Videos of the signs are available at the following URL: https://github.com/CodeCruncher63/ISCR2025

Fig.1: Starting and ending positions for the Italian word amare (to love) in LIS.

"Andare" (To Go). This sign is performed with both arms, which move differently: the left arm remains stationary while the right arm performs a single movement without repetition. The gesture starts as shown in Fig.2a and ends as shown in Fig.2b.

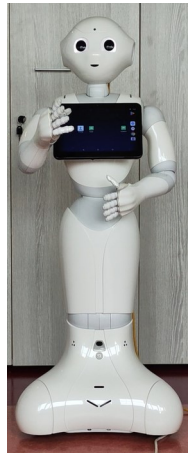
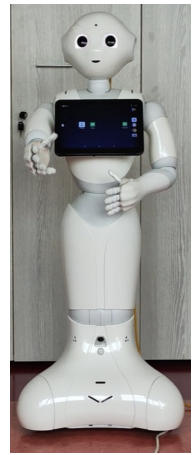

(a) Starting position                                (b) Ending position

Fig.2: Starting and ending positions for the Italian word andare (to go) in LIS.

3.2        Automating the Creation of LIS Signs on Pepper

To streamline the implementation of sign language animations for the Pepper robot, we developed a toolchain that automates the conversion of 3D hand trajectories into joint values and animation files. The system comprises two main components: a MATLAB function that calculates the inverse kinematics for the right arm using a numerical solver, and a Python script that serves as a wrapper, managing user input and generating the final animation file.

We chose to use a numeric solver since inverting the kinematics equations by hand was too complex. Furthermore, the inverted equations that can be found online [29] were obtained by making some simplifications of the inverted kinematics equations that can not be applied in this study.

The MATLAB function receives 5 parameters: the 3D coordinates to be reached, its orientation, and hand configuration (open, closed, or neutral), a flag to mirror the joint values on the left arm and the weights to be used. Then it checks the correctness of the parameters and uses the *inverseKinematics* system object, from the Robotics System Toolbox, to create a numeric solver which computes the joint values. The robot's kinematic model used in the system object was derived from a customized *Unified Robot Description Format* (URDF file) that contains a description of the robot anatomic model; in particular, we considered only the torso and the right arm model description.

The Python script handles data preprocessing and user-defined parameters such as movement timing, solver weights, and the mirroring option. It invokes the MATLAB function passing all the needed inputs and receives as output the joint values. Then it writes the values in the Animation Editor-compatible .qianim XML format, which can be loaded into the Animation Editor (see Fig. 3) and can be used to move the robot in the same way as a file hand-made in this editor.

```xml
<?xml version="1.0" encoding="UTF-8"?>
<Animation xmlns:editor="http://www.ald.softbankrobotics.com/animation/editor" typeVersion="2.0" editor:fps="25">
    <ActuatorCurve fps="25" actuator="LShoulderPitch" mute="false" unit="degree">
        <Key value="90.5273514" frame="0">
            <Tangent side="right" abscissaParam="10" ordinateParam="0" editor:interpType="bezier_auto"/>
        </Key>
        <Key value="90.5273514" frame="30">
            <Tangent side="left" abscissaParam="-10" ordinateParam="0" editor:interpType="bezier_auto"/>
            <Tangent side="right" abscissaParam="6.66666667" ordinateParam="0" editor:interpType="bezier_auto"/>
        </Key>
        <Key value="90.5273514" frame="50">
            <Tangent side="left" abscissaParam="-6.66666667" ordinateParam="0" editor:interpType="bezier_auto"/>
            <Tangent side="right" abscissaParam="6.66666667" ordinateParam="0" editor:interpType="bezier_auto"/>
        </Key>
        <Key value="90.5273514" frame="80">
            <Tangent side="left" abscissaParam="-6.66666667" ordinateParam="0" editor:interpType="bezier_auto"/>
        </Key>
    </ActuatorCurve>
...
```

Fig.3: A fragment of a qianim-XML example

This approach successfully automated the generation of 38 out of the 52 manually implemented signs (73%). Ten signs could not be reproduced due to asymmetric or body-involving movements, and four others failed due to kinematic infeasibility or solver instability. Nonetheless, the method proved advantageous for generating smooth, circular movements and significantly reduced the manual workload. On average, the manual creation of a sign required approximately 45–60 minutes, while the automated method reduced this to under 10 minutes per sign, including coordinate setup and execution.

The main difference between the automated signs and the manually crafted ones, was in the elbow movement: the manually developed signs had the elbow staying still while the arm was moving around it; on the contrary, in the automated signs, the joint values found by the IK solver also involved the elbow moving together with the arm. Anyway, this difference in movement had no impact on the recognition ability of the Deaf student and his sign language interpreter. These results, namely 38 out of the 52, confirm the feasibility of automated sign generation for a substantial subset of LIS signs on Pepper and pave the way for further improvements through solver refinement or data-driven methods.

## 4   Exploratory Evaluation

To assess the intelligibility and clarity of the signs produced by the Pepper robot, we designed an on line evaluation based on a questionnaire targeting sign language users. The aim was to determine whether the robot's gestures, both individual signs and short sentences, could be correctly interpreted by human observers.

The questionnaire consisted of two parts:

- Single-choice closed questions: 15 questions, each showing a video of Pepper performing a single sign. Participants were asked to select the correct sign from four options: three plausible alternatives, one of which was correct, and one "None of the above" option;
- Open-ended questions: 2 questions in which Pepper performed a short sentence composed of multiple signs. Participants were asked to write the meaning of each sentence in natural language. Note that in this experiment, for sake of simplicity, we are neglecting the "spatial accord" among signs of a sentence, a feature that in most sign languages is used to assign (or confirm) the syntactic roles (e.g. *patient*) to the participants [23]. We believe that this simplification so not impact the findings of this exploratory experiment.

### 4.1   Participants

The questionnaire was completed by 12 participants, 4 of whom were LIS interpreters, while the remaining were Deaf individuals who use Italian Sign Language (LIS) on a daily basis. Gender was not collected in order to preserve participant anonymity and focus the analysis solely on sign recognition performance, which was not expected to vary significantly by gender. Participants were between 18 and 60 years old and came from various regions of Italy. Regional differences in LIS usage may have influenced recognition in some cases, although the signs implemented aimed to reflect the most commonly used versions of the signs. Participants were simply instructed to respond sincerely based on their interpretation of the robot's gestures.

### 4.2   Results

Single-Choice Questions The 15 presented signs were chosen among the implemented signs with the help of the LIS interpreter and the Deaf student. The idea was that signs performed with one and two hands were well represented among the chosen signs. Moreover, their intrinsic and grammatical meaning was taken into account and we picked signs belonging to these different categories.

Table 1 summarizes the recognition rates and key observations for each tested sign.

To evaluate whether the recognition of each sign was significantly above chance level (25%), we performed one-tailed binomial tests. The results are summarized in Table 1. Several signs, including *Dimenticare*, *Finire/Fatto*, *Shampoo*, and *Università*, showed statistically significant recognition rates ($p < 0.0001$). Other signs, such as *Doccia* and *Che/Come?*, also yielded recognition rates significantly above chance.

Conversely, signs like *Insegnare* and *Libro* did not reach significance, likely due to articulation constraints.

| Sign | Recognition Rate (%) | Correct (out of 12) | Binomial value p- | Notes |
|---|---|---|---|---|
| Acqua (Water) | 50.0 | 6 | 0.1035 | Often confused with "Apple" (25%); intelligibility confirmed. |
| Antipatico (Unpleasant) | 58.3 | 7 | 0.0436 | Some false recognition (17%). |
| Capriccioso (Capricious) | 58.3 | 7 | 0.0436 | Some false recognition (17%). |
| Che/Come? (What/How?) | 83.3 | 10 | < 0.0001 | High recognition confirms validity. |
| Chiedere (To ask) | 41.7 | 5 | 0.2267 | Mixed responses; confusion with "Insegnare". |
| Dimenticare (To forget) | 100.0 | 12 | < 0.0001 | Iconic and well-executed. |
| Doccia (Shower) | 66.7 | 8 | 0.0041 | Some confusion with "Idea". |
| Finire/Fatto (Done) | 100.0 | 12 | < 0.0001 | Fully recognized. |
| Idea | 91.7 | 11 | < 0.0001 | Widely recognized. |
| Insegnare (To teach) | 33.3 | 4 | 0.4110 | Low recognition, high confusion. |
| Libro (Book) | 50.0 | 6 | 0.1035 | Recognition affected by Pepper's movement limitations. |
| Profumo (Perfume) | 0.0 | 0 | 1.0000 | All selected "Good" due to gesture similarity. |
| Shampoo | 100.0 | 12 | < 0.0001 | Fully recognized. |
| Spaventarsi (To get scared) | 58.3 | 7 | 0.0436 | None of the above answer, 50% selected it correctly. |
| Università (University) | 100.0 | 12 | < 0.0001 | Fully recognized. |

Table 1: Recognition accuracy, binomial test results, and qualitative notes for each sign.

Open-Ended Questions Performance in the open-ended section was notably lower than in the single-choice section, suggesting greater difficulty in interpreting full signed sentences:

– Sentence 1: *"Ho mangiato una mela"* (I ate an apple), composed of *mela* (apple) – *mangiare* (to eat) – *fatto* (done).
  Most participants interpreted the sequence as *"Ho bevuto e mangiato"* (I drank and ate), likely due to the confusion between the signs for "apple" and "water". Only one participant correctly interpreted the full sentence. All other 11

- participants attempted a response and correctly recognized at least 2 signs out of 3.
- Sentence 2: *"Vado a casa a studiare"* (I go home to study), composed of *casa* (home) – *studiare* (to study) – *andare* (to go).
  This sentence had a low recognition rate. One participant gave a correct response; one provided a nearly correct interpretation. The 11 remaining participants either misunderstood the sentence or declared it unintelligible. The sign for *casa* was likely a limiting factor due to Pepper's restricted movement range.

To analyze more in depth the open-ended responses, we applied an inductive thematic coding approach [7]. Three main categories emerged: (i) lexical confusion due to similar handshape/movement (e.g., mela vs. acqua), (ii) sentence structure misinterpretation, and (iii) partial recognition with correct identification of individual signs but incorrect sequencing. For example, one participant interpreted the sentence 'mela – mangiare – fatto' as 'ho bevuto e mangiato,' reflecting confusion between 'apple' and 'water,' likely due to hand orientation limits. This suggests that iconic similarity can hinder clarity when physical constraints distort precise configurations.

## 5   Discussion and Conclusions

The evaluation results confirm that Pepper is capable of performing a substantial number of LIS signs with a good level of intelligibility. Recognition accuracy was notably higher for individual signs than for complete sentences, likely due to the added complexity of temporal sequencing and the cognitive load involved in integrating multiple gestures into a coherent semantic unit. Performance varied significantly across signs, some signs were occasionally misinterpreted, often due to Pepper's physical constraints.

To answer research question RQ1, the study identified several technical and communicative limitations that constrain the use of a commercial social robot such as Pepper for expressing LIS in accessible Deaf–robot interaction.

From a technical standpoint, the most prominent limitation is Pepper's inability to move fingers independently, which excludes a wide range of handshapes fundamental to LIS. Additionally, constrained wrist and elbow mobility, along with the presence of the chest-mounted tablet, interferes with gesture execution, particularly for signs that require close-body movements or fine spatial articulation. These constraints directly impact the robot's ability to faithfully reproduce the complexity of LIS signs.

From a communicative perspective, while many isolated signs were recognized correctly by LIS users, especially those with iconic or simplified motion—sentencelevel comprehension proved significantly more challenging. This highlights Pepper's limited capacity to convey temporal and syntactic structure, which is essential for interpreting full utterances in sign language. Furthermore, some signs were misinterpreted due to articulation ambiguity, while other signs were unexpectedly recognized, suggesting a nuanced relationship between physical execution and human perception of signs.

While Pepper demonstrates the potential to perform a limited yet meaningful subset of LIS signs, its current embodiment imposes notable constraints on expressivity and intelligibility. These findings underscore the importance of integrating multimodal channels (e.g., visual display, expressive avatars) and maintaining a participatory design process to enhance accessibility in future human–robot interaction systems.

As limitation, we acknowledge that the evaluation was conducted with a relatively small sample of 12 participants. While this number is within acceptable bounds for exploratory user studies in HRI, it does limit the statistical power and generalizability of the findings. However, it is important to note that recruiting Deaf participants for this type of research presents some challenge. Despite our efforts to collaborate with local communities and institutions, we encountered a degree of hesitation and resistance from potential participants. These barriers may reflect broader issues of trust. As such, we consider the current sample size not only as a methodological limitation, but also as a reflection of the practical and ethical complexity of engaging marginalized groups in HRI research.

Future work should aim to foster longer-term collaborations with Deaf organizations and educators, build trust through participatory approaches, and develop more inclusive research environments to support broader recruitment and sustained engagement.

Although this study focused on physical sign reproduction, future work may explore multimodal augmentation. The project LIS4ALL has showed the advantages of multimodality in LIS translation. The main idea was to augment a virtual agent with "blended written forms", which are a sort of road signs, for communicate the names of less-known rail stations [14,22]. In the case of Pepper, multimodality could consist in the coordination of arms and images/videos. For instance, integrating Pepper's tablet to display a LIS avatar (or subtitles) could enhance intelligibility for complex signs. However, such integration requires careful synchronization and user validation, particularly in real-time contexts. We consider this a promising yet technically demanding direction for future participatory co-design efforts.

Beyond experimental contexts, future applications of this technology could include interactive environments such as museums, public information points, or healthcare facilities, where Deaf users may come into contact with service robots. In these scenarios, even a partial command of LIS, supported by multimodal output, could play a key role in reducing communication barriers and fostering more inclusive interactions. Advancing in this direction will require ongoing collaboration with the Deaf community, user-centered design processes, and testing in real-world conditions to evaluate long-term usability and acceptance.

**Acknowledgments.** We are particularly grateful to Flavio and his LIS interpreter Simonetta for the considerable help in choosing which signs to implement, generate them and discard the ones that Pepper was not performing correctly.